\documentclass[11pt,a4paper]{article}
\usepackage[hyperref]{naaclhlt2019}
\usepackage{times}
\usepackage{latexsym}
\usepackage{graphicx}
\usepackage{multirow}
\usepackage{booktabs}
\usepackage{amsmath}
\usepackage{amssymb}
\usepackage{csquotes}
\usepackage{url}
\usepackage[normalem]{ulem}

\usepackage[textsize=scriptsize]{todonotes}
\definecolor{blue}{RGB}{0, 93, 170}

\aclfinalcopy 

\title{Imposing Label-Relational Inductive Bias\\ for Extremely Fine-Grained Entity Typing}

\author{
 Wenhan Xiong$^\dagger$,
 Jiawei Wu$^\dagger$,
 Deren Lei$^\dagger$,
 Mo Yu$^\ast$, 
 Shiyu Chang$^\ast$, 
 Xiaoxiao Guo$^\ast$, 
 William Yang Wang$^\dagger$
\\ 
 $^\dagger$ University of California, Santa Barbara\\
 $^\ast$ IBM Research\\
 \{xwhan, william\}@cs.ucsb.edu, yum@us.ibm.com, \{shiyu.chang, xiaoxiao.guo\}@ibm.com  
 }

\date{}

\begin{document}
\maketitle
\begin{abstract}


Existing entity typing systems usually exploit the type hierarchy provided by knowledge base (KB) schema to model label correlations and thus improve the overall performance.   Such techniques, however, are not directly applicable to more open and practical scenarios where the type set is not restricted by KB schema and includes a vast number of free-form types. To model the underlying label correlations without access to manually annotated label structures, we introduce a novel label-relational inductive bias, represented by a graph propagation layer that effectively encodes both global label co-occurrence statistics and word-level similarities. On a large dataset with over 10,000 free-form types, the graph-enhanced model equipped with an attention-based matching module is able to achieve a much higher recall score while maintaining a high-level precision.  Specifically, it achieves a $15.3\%$ relative F1 improvement and also less inconsistency in the outputs.  We further show that a simple modification of our proposed graph layer can also improve the performance on a conventional and widely-tested dataset that only includes KB-schema types.\footnote{\url{https://github.com/xwhan/Extremely-Fine-Grained-Entity-Typing}}

\end{abstract}

\section{Introduction}

\begin{table}[t]
\centering
\begin{tabular}{p{4.8cm}|p{2cm}}
\toprule
\bf Context & \bf Types \\
\midrule
\multirow{5}{4.9cm}{\small{\textbf{Big Show} then appeared at One Night Stand, attacking Tajiri, Super Crazy, and the Full Blooded Italians after their tag team match}} &  \multirow{2}{2cm}{\small{\textcolor{red}{person$^\dagger$}, \textcolor{blue}{television\_program$^\star$}}} \\ 
& \\\cline{2-2}
& \multirow{3}{2cm}{\small{\textit{person, athlete, wrestler, entertainer}}}\\
& \\
& \\
\midrule
\multirow{4}{4.9cm}{\small{The women‘s pole vault at the 2010 IAAF World Indoor Championships was held at the ASPIRE Dome on 12 and 14 \textbf{March}.}} & \multirow{2}{2cm}{\small{\textcolor{red}{month$^\dagger$}, \textcolor{blue}{event$^\star$}}}\\ 
& \\
\cline{2-2}
& \multirow{2}{2cm}{\small{\textit{date, month}}}\\
& \\
\bottomrule
\end{tabular}
\caption{Examples of inconsistent predictions produced by existing entity typing system that does not model label correlations. We use different subscript symbols to indicate contradictory type pairs and show the  ground-truth types in \textit{italics}.}
\label{tab:example}
\end{table}

Fine-grained entity typing is the task of identifying specific semantic types of entity mentions in given contexts. In contrast to general entity types (\emph{e.g.}, organization, event), fine-grained types (\emph{e.g.}, political party, natural disaster) are often more informative and can provide valuable prior knowledge for a wide range of NLP tasks, such as coreference resolution~\cite{durrett2014joint}, relation extraction~\cite{yaghoobzadeh2016noise} and question answering~\cite{lee2006fine,yavuz2016improving}.

In practical scenarios, a key challenge of entity typing is to correctly predict \textit{multiple} ground-truth type labels from a large candidate set that covers a wide range of types in different granularities. In this sense, it is essential for models to effectively capture the inter-label correlations. For instance, if an entity is identified as a ``criminal", then the entity must also be a ``person",  but it is less likely for this entity to be a ``police officer" at the same time. When ignoring such correlations and considering each type separately, models are often inferior in performance and prone to inconsistent predictions. As shown in Table~\ref{tab:example}, an existing model that independently predicts different types fails to reject predictions that include apparent contradictions. 

Existing entity typing research often address this aspect by explicitly utilizing a given type hierarchy to design hierarchy-aware loss functions~\cite{ren2016label,xu2018neural} or enhanced type label encodings~\cite{shimaoka2016neural} that enable parameter sharing between related types. These methods rely on the assumption that the underlying type structures are predefined in entity typing datasets. For benchmarks annotated with the knowledge base (KB) guided distant supervision, this assumption is often valid since all types are from KB ontologies and naturally follow tree-like structures. However, since knowledge bases are inherently incomplete~\cite{min2013distant}, existing KBs only include a limited set of entity types. Thus, models trained on these datasets fail to generalize to lots of unseen types. In this work, we investigate entity typing in a more open scenario where the type set is not restricted by KB schema and includes over 10,000 free-form types~\cite{choi2018ultra}. As most of the types do not follow any predefined structures, methods that explicitly incorporate type hierarchies cannot be straightforwardly applied here.

To effectively capture the underlying label correlations without access to known type structures, we propose a novel label-relational inductive bias, represented by a graph propagation layer that operates in the latent label space. Specifically, this layer learns to incorporate a label affinity matrix derived from global type co-occurrence statistics and word-level type similarities. It can be seamlessly coupled with existing models and jointly updated with other model parameters. Empirically, on the Ultra-Fine dataset~\cite{choi2018ultra}, the graph layer alone can provide a significant $11.9\%$ relative F1 improvement over previous models. Additionally, we show that the results can be further improved ($11.9\%$ $\rightarrow$ $15.3\%$) with an attention-based mention-context matching module that better handles \textit{pronouns} entity mentions. With a simple modification, we demonstrate that the proposed graph layer is also beneficial to the widely used OntoNotes dataset, despite the fact that samples in OntoNotes have lower label multiplicity (\emph{i.e.}, average number of ground-truth types for each sample) and thus require less label-dependency modeling than the Ultra-Fine dataset. 

To summarize, our major contribution includes:
\begin{itemize}
    \item We impose an effective label-relational bias on entity typing models with an easy-to-implement graph propagation layer, which allows the model to implicitly capture type dependencies;
    \item We augment our graph-enhanced model with an attention-based matching module, which constructs stronger interactions between the mention and context representations;
    \item Empirically, our model is able to offer significant improvements over previous models on the Ultra-Fine dataset and also reduces the cases of inconsistent type predictions.
\end{itemize}

\section{Related Work}
\paragraph{Fine-Grained Entity Typing} The task of fine-grained entity typing was first thoroughly investigated in \cite{ling2012fine}, which utilized Freebase-guided distant supervision (DS)~\cite{mintz2009distant} for entity typing and created one of the early large-scale datasets. Although DS provides an efficient way to annotate training data, later work~\cite{gillick2014context} pointed out that entity type labels induced by DS ignore entities' local context and may have limited usage in context-aware applications. Most of the following research has since focused on testing in context-dependent scenarios. While early methods~\cite{gillick2014context,yogatama2015embedding} on this task rely on well-designed loss functions and a suite of hand-craft features that represent both context and entities,~\citet{shimaoka2016attentive} proposed the first attentive neural model which outperformed feature-based methods with a simple cross-entropy loss.

\paragraph{Modeling Entity Type Correlations} To better capture the underlying label correlations,~\citet{shimaoka2016neural} employed a hierarchical label encoding method and AFET~\cite{ren2016afet} used the predefined label hierarchy to identify noisy annotations and proposed a partial-label loss to reduce such noise. A recent work~\cite{xu2018neural} proposed hierarchical loss normalization which alleviated the noise of too specific types. Our work differs from these works in that we do not rely on known label structures and aim to learn the underlying correlations from data. \citet{rabinovich2017fine} recently proposed a structure-prediction approach which used type correlation features. The inference on their learned factor graph is approximated by a greedy decoding algorithm, which outperformed unstructured methods on their own dataset. Instead of using an explicit graphical model, we enforce a relational bias on model parameters, which does not introduce extra burden on label decoding.

\begin{figure*}[t]
\centering
\includegraphics[width=1.0\linewidth]{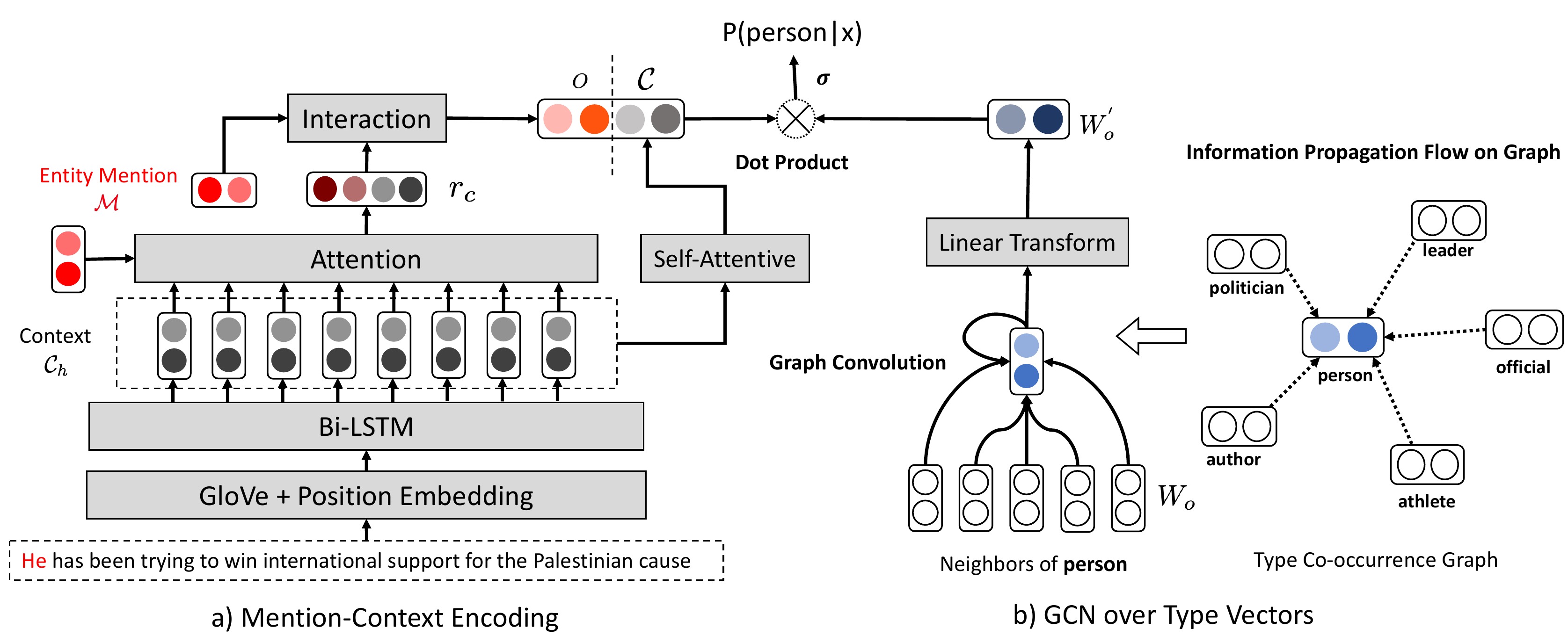}
\caption{Overview of the process to make predictions on the type \textbf{``person"}. \textbf{a)} Modules used to extract mention and context aware representations. \textbf{b)} An illustration of the graph layer operating over the type vector of \textbf{``person"}.}
\label{model}
\end{figure*}

\section{Task Definition}
Specifically, the task we consider takes a raw sentence $C$ as well as an entity mention span $M$ inside $C$ as inputs, and aims to predict the correct type labels $T_{m}$ of $M$ from a candidate type set $\mathcal{T}$, which includes more than 10,000 free-form types. The entity span $M$ here can be named entities, nominals and also pronouns. The ground-truth type set $T_{m}$ here usually includes more than one types (approximately five types on average), making this task a multi-label classification problem.

\section{Methodology}
In this section, we first briefly introduce the neural architecture to encode raw text inputs. Then we describe the matching module we use to enhance the interaction between the mention span and the context sentence. Finally, we move to the label decoder, on which we impose the label-relational bias with a graph propagation layer that encodes type co-occurrence statistics and word-level similarities. Figure~\ref{model} provides a graphical overview of our model, with \ref{model}a) illustrating both the text encoders and the matching module, and \ref{model}b) showing an example of graph propagation.

\subsection{Representation Model}
Our base model to encode the context and the mention span follows existing neural approaches~\cite{shimaoka2016attentive,xu2018neural,choi2018ultra}. To encode the context, we first apply a standard Bi-LSTM, which takes GloVe~\cite{pennington2014glove} embeddings and position embeddings (three vectors representing positions before, inside or after the mention span) as inputs and outputs the hidden states at each time step $t\in[1,l_c]$. With the derived hidden states $\mathcal{C}_h \in \mathbb{R}^{l_c \times h_c}$, we then apply a self-attentive encoder~\cite{mccann2017learned} on the top to get the final context representation $\mathcal{C}$. For the entity mention span, we concatenate the features derived by a character-level CNN and a similar self-attentive encoder. We denote the final mention representation as $\mathcal{M}$.\footnote{Please refer to \cite{shimaoka2016neural} and \cite{choi2018ultra} for more detailed descriptions.} 

\subsection{Mention-Context Interaction}
\label{inter}
Since most previous datasets only consider \textit{named entities}, a simple concatenation of the two features $[\mathcal{C}; \mathcal{M}]$ followed by a linear output layer~\cite{shimaoka2016attentive,shimaoka2016neural} usually works reasonably well when making predictions. This suggests that $\mathcal{M}$ itself provides important information for recognizing entity types. However, as in our target dataset, a large portion of entity mentions are actually \textit{pronouns}, such as ``he'' or ``it'', this kind of mentions alone provide only limited clues about general entity types (\emph{e.g.}, ``he" is a ``person") but little information about fine-grained types. In this case, directly appending representation of pronouns does not provide extra useful information for making fine-grained predictions. Thus, instead of using the concatenation operator, we propose to construct a stronger interaction between the mention and context with an attention-based matching module, which has shown its effectiveness in recent natural language inference models~\cite{P16-2022,P17-1152}. 

Formally consider the mention representation $\mathcal{M} \in \mathbb{R}^{h_m}$ and context's hidden feature $\mathcal{C}_h \in \mathbb{R}^{l_c \times h_c}$, where $l_c$ indicates the number of tokens in the context sentence and $h_m, h_c$ denote feature dimensions. We first project the mention feature $\mathcal{M}$ into the same dimension space as $\mathcal{C}_h$ with a linear layer ($W_1 \in \mathbb{R}^{h_m \times h_c}$) and a $tanh$ function\footnote{$tanh$ here is used to make $m_{proj}$ in the same scale as $\mathcal{C}_h$, which was the output of a $tanh$ function inside LSTM.}:
\begin{equation}
    m_{proj} =  tanh(W_1^{T}\mathcal{M}),
\end{equation} then we perform bilinear attention matching between $m_{proj}$ and $\mathcal{C}_h$, resulting in an affinity matrix $\mathcal{A}$ with dimension $\mathcal{A} \in \mathbb{R}^{1 \times l_c}$:
\begin{equation}
    \mathcal{A} = m_{proj} \times W_a \times \mathcal{C}_h,
\end{equation}
where $W_a \in\mathbb{R}^{h_c\times h_c}$ is a learnable matrix. If we consider the mention feature as query and the context as memory, we can use the affinity matrix to retrieve the relevant parts in the context:
\begin{align}
    \bar{\mathcal{A}} &= softmax(\mathcal{A}) \\
    r_c &= \bar{\mathcal{A}} \times \mathcal{C}_h.
\end{align}
With the projected mention representation $m_{proj}$ and the retrieved context feature $r_c$, we define the following interaction operators:
\begin{align}
    r &=  \rho(W_r[r_c; m_{proj}; r_c - m_{proj}])\\
    g &=  \sigma(W_g[r_c; m_{proj}; r_c - m_{proj}]) \\
    o &= g * r + (1-g) * m_{proj},
\end{align}
where $\rho(\cdot)$ is a gaussian error linear unit~\cite{gelu} and $r$ is the fused context-mention feature; $\sigma(\cdot)$ indicates a \textit{sigmoid} function and $g$ is the resulting gating function, which controls how much information in mention span itself should be passed down. We expect the model to focus less on the mention representation when it is not informative. The concatenation $[r_c; m_{proj}; r_c - m_{proj}]$ here is supposed to capture different aspects of the interactions. To emphasize the context's impact, we finally concatenate the extracted context feature ($\mathcal{C}$) with the output ($o$) of the matching module ($f = [o;\mathcal{C}]$) for prediction.

\subsection{Imposing Label-Relational Inductive Bias}
For approaches that ignore the underlying label correlations, the type predictions are considered as $N$ independent binary classification problems, with $N$ being the number of types. If we denote the feature extracted by any arbitrary neural model as $f\in \mathbb{R}^{d_f}$, then the probability of being any given type is calculated by:
\begin{equation}
    p = \sigma(W_o f), W_o \in \mathbb{R}^{N \times d_f}.\label{eqn:classifier}
\end{equation}
We can see that every row vector of $W_o$ is responsible for predicting the probability of one particular type. We will refer the row vectors as type vectors for the rest of this paper. As these type vectors are independent, the label correlations are only implicitly captured by sharing the model parameters that are used to extract $f$. We argue that the paradigm of parameter sharing is not enough to impose strong label dependencies and the values of type vectors should be better constrained. 

A straightforward way to impose the desired constraints is to add extra regularization terms on $W_o$. We first tested several auxiliary loss functions based on the heuristics from GloVe~\cite{pennington2014glove}, which operates on the type co-occurrence matrix. However, the auxiliary losses only offer trivial improvements in our experiments. Instead, we find that directly imposing a model-level inductive bias on the type vectors turns out to be a more principled solution. This is done by adding a graph propagation layer over randomly initialized $W_o$ and generating the updated type vectors $W^{'}_o$, which is used for final prediction. Both $W_o$ and the graph convolution layer are learned together with other model parameters. We view this layer as the key component of our model and use the rest of this section to describe how we create the label graph and compute the propagation over the graph edges.

\paragraph{Label Graph Construction}
In KB-supervised datasets, the entity types are usually arranged in tree-like structures. Without any prior about type structures, we consider a more general graph-like structure. While the nodes in the graph straightforwardly represent entity types, the meaning of the edges is relatively vague, and the connections are also unknown. In order to create meaningful edges using training data as the only resource, we utilize the type co-occurrence matrix: if two type $t_1$ and $t_2$ both appear to be the true types of a particular entity mention, we will add an edge between them. In other words, we are using the co-occurrence statistics to approximate the pair-wise dependencies and the co-occurrence matrix now serves as the adjacent matrix. Intuitively, if $t_2$ co-appears with $t_1$ more often than another type $t_3$, the probabilities of $t_1$ and $t_2$ should have stronger dependencies and the corresponding type vectors should be more similar in the vector space. In this sense, we expect each type vector to effectively capture the local neighbor structure on the graph.

\paragraph{Correlation Encoding via Graph Convolution}
To encode the neighbor information into each node's representation, we follow the propagation rule defined in Graph Convolution Network (GCN)~\cite{gcn}. In particular, with the adjacent or co-occurrence matrix $A$, we define the following propagation rule on $W_o$:
\begin{align}
    W_o^{'} &= \tilde{D}^{-\frac{1}{2}}\tilde{A}\tilde{D}^{-\frac{1}{2}}W_o T\\
    \tilde{A} &= A + I_N.
\end{align}
Here $T \in \mathbb{R}^{d_f \times d_f}$ is the transformation matrix and $I_N$ is an identity matrix used to add self-connected edges. $\tilde{D}$ is a diagonal degree matrix with $\tilde{D}_{ii} = \sum_j \tilde{A}_{ij}$, which is used to normalize the feature vectors such that the number of neighbors does not affect the scale of transformed feature vectors. In our experiments, we find that an alternative propagation rule
\begin{equation}
    W_o^{'} = \tilde{D}^{-1}\tilde{A} W_o T
\end{equation}
works similarly well and is more efficient as it involves less matrix multiplications. If we look closely and take each node out, the propagation can be written as
\begin{equation}
    \label{prop}
    W_o^{'}[i,:] = \frac{1}{\sum_j \tilde{A}_{ij}} (\sum_j \tilde{A}_{ij}W_o[j,:]T).
\end{equation}
From this formula, we can see that the propagation is essentially gathering features from the first-order neighbors. In this way, the prediction on type $t_i$ is dependent on its neighbor types. 
\begin{figure}[t]
\centering
\includegraphics[width=0.9\linewidth]{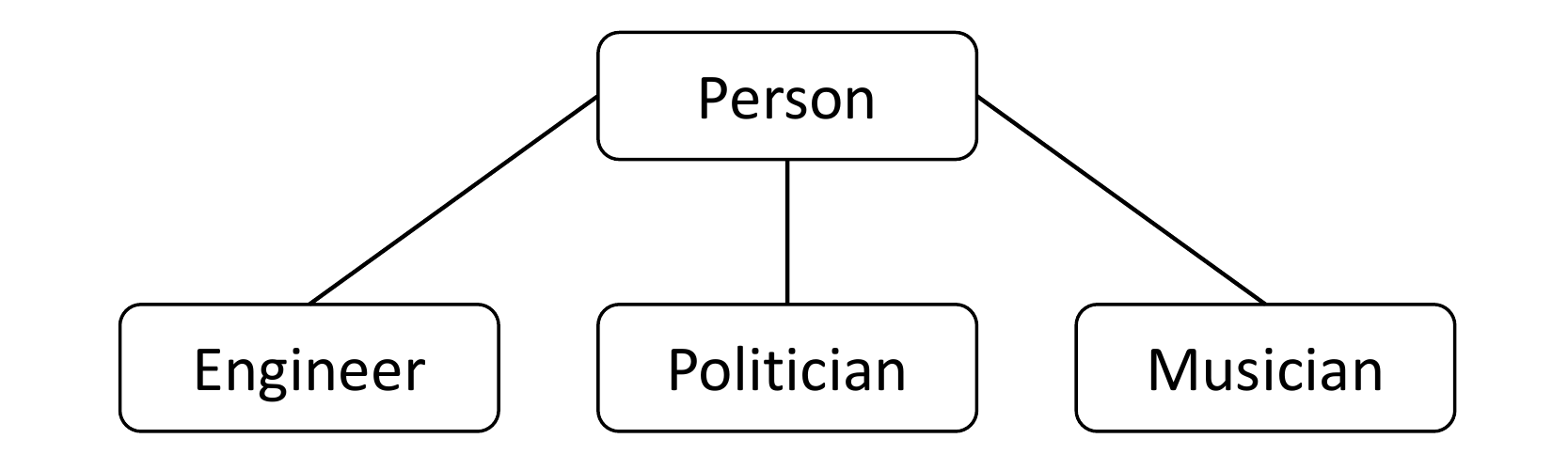}
\caption{A snippet of the underlying type co-occurrence graph. Multiple edges between nodes are omitted here for clarity.}
\label{label}
\end{figure}

Compared to original GCNs that often use multi-hop propagations (\emph{i.e.}, multiple graph layers connected by nonlinear functions) to capture higher-order neighbor structures. We only apply one-hop propagation and argue that high-order label dependency is not necessarily beneficial in our scenario and might introduce false bias. A simple illustration is shown in Figure~\ref{label}. We can see that propagating 2-hop information introduces undesired inductive bias, since types that are more than 1-hop away (\emph{e.g.}, ``Engineer" and ``Politician") usually do not have any dependencies. In fact, some of the 2-hop type pairs can be contradictory types (\emph{e.g.}, ``police" and ``prisoner"). This hypothesis is consistent with our experiment results: adding more than one graph layer leads to worse results. Additionally, we also omit GCN's nonlinear activation which introduces unnecessary constraints on the scale of $W_o^{'}$, with which we calculate the unscaled scores before calculating the probability via a sigmoid function.

\subsection{Leveraging Label Word Embeddings}
As the type labels are all written as text phrases, an interesting question is whether we can exploit the semantics provided by pre-trained word embeddings to improve entity typing. We explore this possibility by using the cosine similarity of word embeddings. We first calculate type embeddings by simply summing the embeddings of all tokens in the type name. Then we build a label affinity matrix $A_{word}$ by calculating pair-wise cosine similarities. With the assumption that word-level similarity measures some degree of label dependency, we propose to integrate $A_{word}$ into the graph convolution layer following
\begin{align}
\label{scale}
    A^{'}_{word} &= (A_{word} + 1) / 2 \\ 
    W_o^{'} &= \tilde{D}^{-1}(\tilde{A} + \lambda  A^{'}_{word})W_o T.
\end{align}
Here Equation~\ref{scale} scales the similarity value into $(0,1]$ to avoid negative edge weights, which might introduce numerical issues when calculating $\tilde{D}^{-1}$. $\lambda$ is a trainable parameter used to weight the impact of word-level similarities. As will be shown in Section~\ref{exp}, this simple augmentation provides further improvement over our original model.

\section{Experiments}
\label{exp}
\subsection{Experiment Setup}

\begin{figure}[t]
\centering
\includegraphics[width=0.95\linewidth]{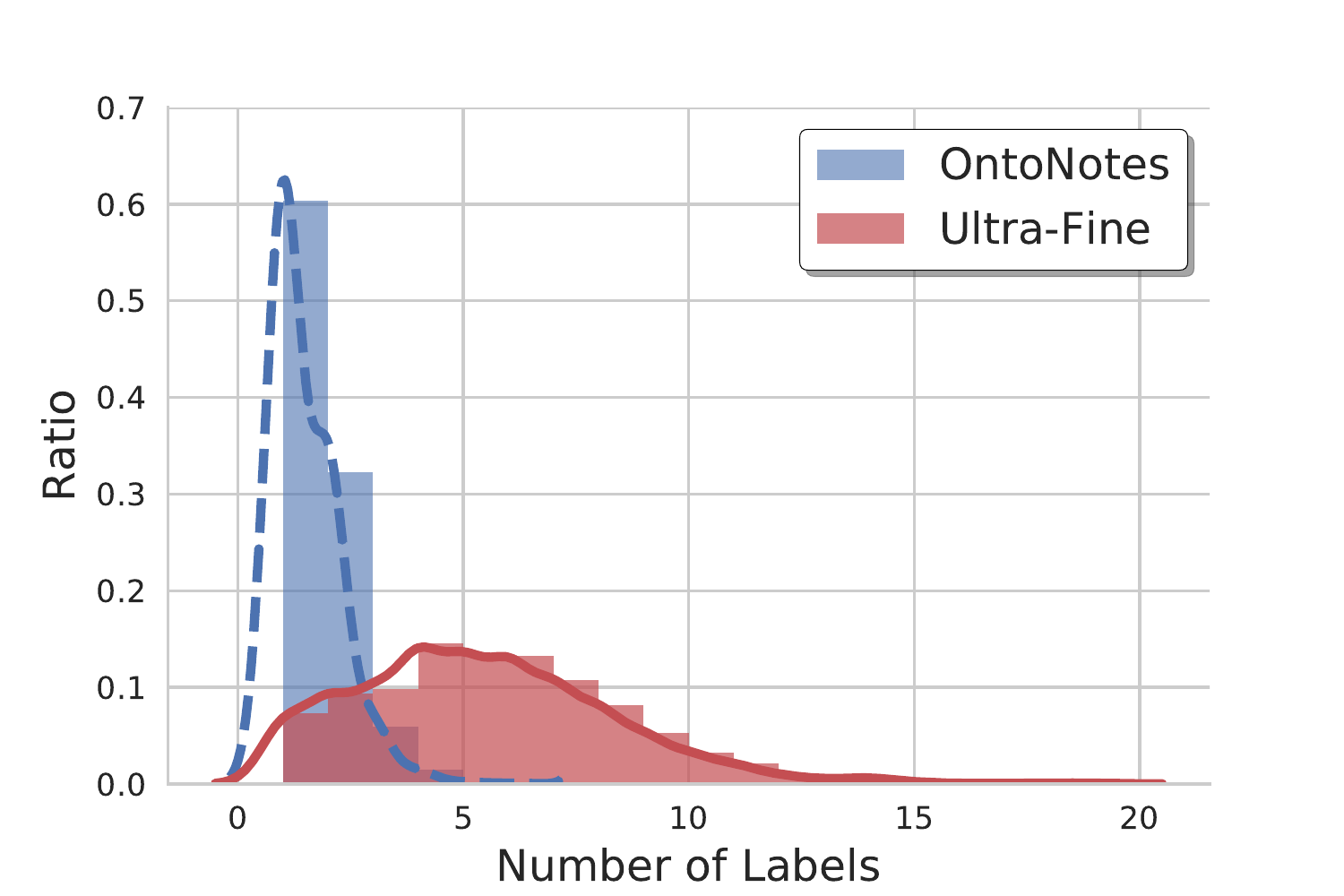}
\caption{Label multiplicity distribution of the datasets.}
\label{dataset}
\end{figure}

\paragraph{Datasets} Our experiments mainly focus on the Ultra-Fine entity typing dataset which has 10,331 labels and most of them are defined as free-form text phrases. The training set is annotated with heterogeneous supervisions based on KB, Wikipedia and head words in dependency trees, resulting in about 25.2M\footnote{\citet{choi2018ultra} use the licensed Gigaword to build part of the dataset, while in our experiments we only use the open-sourced training set which has approximately 6M training samples.} training samples. This dataset also includes around 6,000 crowdsourced samples. Each of these samples has five ground-truth labels on average. For a fair comparison, we use the original test split of the crowdsourced data for evaluation. To better understand the capability of our model, we also test our model on the commonly-used OntoNotes~\cite{gillick2014context} benchmark. It is worth noting that this dataset is much smaller and has lower label multiplicity than the Ultra-Fine dataset, \emph{i.e.}, each sample only has around 1.5 labels on average. Figure~\ref{dataset} shows a comparison of these two datasets.

\paragraph{Baselines} For the Ultra-Fine dataset, we compare our model with AttentiveNER~\cite{shimaoka2016attentive} and the multi-task model proposed with the Ultra-Fine dataset. Note that \emph{other models that require pre-defined type hierarchy are not applicable to this dataset}. For experiments on OntoNotes, in addition to the two neural baselines for Ultra-Fine, we compare with several existing methods that explicitly utilize the pre-defined type structures in loss functions. Namely, these methods are AFET~\cite{ren2016afet}, LNR~\cite{ren2016label} and NFETC~\cite{xu2018neural}.

\paragraph{Evaluation Metrics} On Ultra-Fine, we first evaluate the mean reciprocal rank (MRR), macro precision(P), recall (R) and F1 following existing research. As P, R and F1 all depend on a chosen threshold on probabilities, we also consider a more transparent comparison using precision-recall curves. On OntoNotes, we use the standard metrics used by baseline models: accuracy, macro, and micro F1 scores.

\paragraph{Implementation Details} Most of the model hyperparameters, such as embedding dimensions, learning rate, batch size, dropout ratios on context and mention representations are consistent with existing models. Since the mention-context matching module brings more parameters, we apply a dropout layer over the extracted feature $f$ to avoid overfitting. We list all the hyperparameters in the appendix. Models for OntoNotes are trained with standard binary cross-entropy (BCE) losses defined on all candidate labels. When training on Ultra-Fine, we adopt the multi-task loss proposed in \citet{choi2018ultra} which divides the cross-entropy loss into three separate losses over different type granularities. The multi-task objective avoids penalizing false negative types and can achieve higher recalls.

\begin{table*}[t]
\centering
\small
\begin{tabular}{l|cccc|cccc}
\toprule
& \multicolumn{4}{c|}{\textbf{Dev}} & \multicolumn{4}{c}{\textbf{Test}}\\
\cmidrule{2-5}  \cmidrule{6-9} 
\textbf{Model} & \textbf{MRR} & \textbf{P} & \textbf{R} & \textbf{F1} & \textbf{MRR} & \textbf{P} & \textbf{R} & \textbf{F1}  \\ \midrule
AttentiveNER & 0.221 & 53.7 & 15.0 & 23.5 & 0.223 & 54.2 & 15.2 & 23.7 \\
\citet{choi2018ultra} & 0.229 & 48.1 & 23.2 & 31.3 & 0.234 & 47.1 & 24.2 & 32.0 \\ \midrule
\textsc{LabelGCN} & \textbf{0.250} & 50.5 & \textbf{28.7} & \textbf{36.6} & \textbf{0.253} & 50.3 & \textbf{29.2} & \textbf{36.9} \\ \multicolumn{1}{l|}{- w/o word embedding} & 0.245 & 49.4 & 27.8 & 35.6 & 0.249 & 48.7 & 28.3 & 35.8 \\
\multicolumn{1}{l|}{- w/o gcn propagation} & 0.231 & 47.8 & 25.7 & 33.5 & 0.239 & 45.4 & 25.8 & 32.9 \\ 
\multicolumn{1}{l|}{- w/o mention-context interaction} & 0.249 & 53.2 & 25.0 & 34.0 & 0.253 & 54.3 & 25.8 & 35.0 \\
\textsc{LabelGCN} + \emph{threshold-tuning} & \textbf{0.250} & \textbf{55.6} & 25.4 & 35.0 & \textbf{0.253} & \textbf{54.8} & 25.9 & 35.1\\
\bottomrule
\end{tabular}
\caption{Comparison with baseline models on the Ultra-Fine dataset. \emph{Threshold-tuning} gives better performance on all metrics compared to both baselines.}
\label{results}
\vspace{-0.2in}
\end{table*}

\subsection{Evaluation on the Ultra-Fine Dataset}
We report the results on Ultra-Fine in Table~\ref{results}. It is worth mentioning that our model, denoted as \textsc{LabelGCN}, is trained using the unlicensed training set which is smaller than the one used by compared baselines. Even though our model significantly outperforms the baselines, for a fair comparison, we first test our model using the same decision threshold (0.5) used by previous models. In terms of F1, our best model (\textsc{LabelGCN}) outperforms existing methods by a large margin. Compared to \citet{choi2018ultra}, our model improves on both precision and recall significantly. Compared to the AttentiveNER trained with standard BCE loss, our model achieves much higher recall but performs worse in precision. This is due to the fact that when trained with BCE loss, the model usually retrieves only one label per sample and these types are mostly general types\footnote{According to the results of our own implementation of BCE-trained model which achieves similar performance as AttentiveNER.} which are easier to predict. With higher recalls or more retrieved types, achieving high precision requires being accurate on fine-grained types, which are often harder to predict. 

\begin{figure}[t]
\centering
\includegraphics[width=1.0\linewidth]{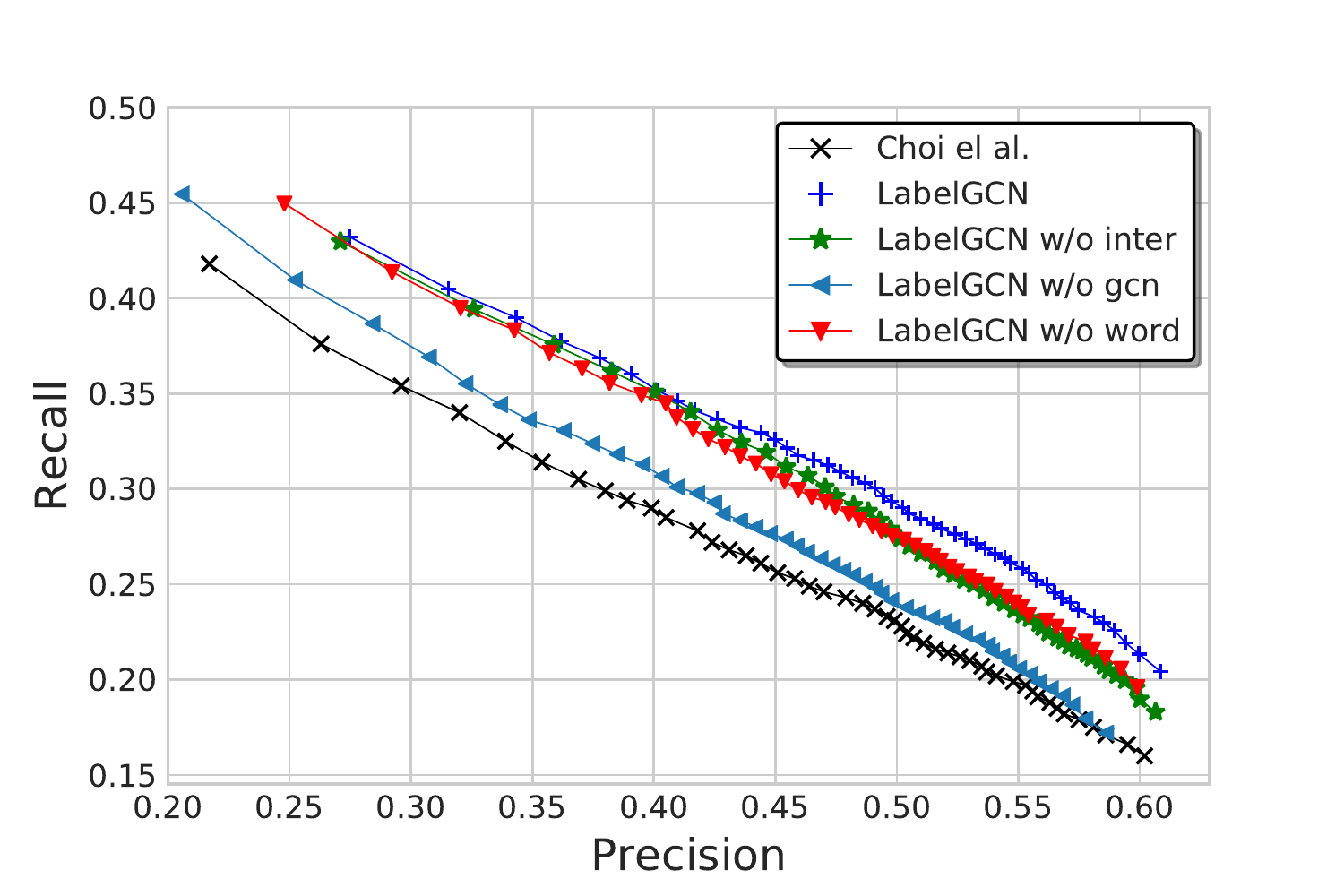}
\caption{Precision-recall curves on Ultra-Fine. The trivial point derived by threshold 0 is omitted here.}
\label{pr}
\end{figure}

\begin{table}[t]
    \centering
    \small
    \begin{tabular}{c|cc}
    \toprule
        \textbf{Model} & \textbf{F1-pronouns} & \textbf{F1-else}\\
        \citet{choi2018ultra} & 35.8 & 32.0 \\ 
        \citet{choi2018ultra} + inter & 38.2 ($\uparrow 2.4$) & 32.8 \\
        \midrule
        \textsc{LabelGCN} w/o inter & 38.6 & 36.8 \\
        \textsc{LabelGCN} & 39.3 ($\uparrow 0.7$) & 36.5\\
        \bottomrule
    \end{tabular}
    \caption{Decomposed validation performance on pronouns and the other entities. Each entry is obtained using the best threshold among the 50 equal-interval thresholds. The corresponding PR curves can be found in the appendix (Figure~\ref{fig:more}).}
    \label{tab:interaction}
    \vspace{-0.1in}
\end{table}

As the precision and recall scores both rely on the decision threshold, different models or different metrics can have different optimal thresholds. As shown by the ``\textsc{LabelGCN} + thresh tuning" entry in Table~\ref{results}, with threshold tuning, our model beats baselines in all metrics. We also see that recall is usually lagging behind precision on this dataset, indicating that F1 score is mainly affected by the recall and tuning towards recall can usually lead to higher F1 scores. For more transparent comparisons, we show the precision-recall curves in Figure~\ref{pr}. These data points are based on the validation performance given by 50 equal-interval thresholds between 0 and 1. We can see there is a clear margin between our model and the multi-task baseline method (LabelGCN vs Choi et al.).

\subsection{Ablation Studies}
To quantify the effect of different model components, we report the performance of model variants in Table~\ref{results} and Figure~\ref{pr}. We can clearly see that the graph convolution layer is the most essential component. The information provided by word embedding is useful and can further improve both precision and recall.
Although Table~\ref{results} seems to indicate the interaction module decreases the precision, we can see from Figure~\ref{pr} that with a proper threshold, the enhanced interaction actually improves both precision and recall.
In term of this, we recommend future research to use PR curves for more accurate model analysis.

\subsection{Fine-Grained Performance for Pronouns}
As discussed in Section~\ref{inter}, the mention representation of pronouns provide limited information about fine-grained types. We investigate the effect of the enhanced mention-context interaction by analyzing the decomposed performance on pronouns and other kinds of entities. From the results in Table~\ref{tab:interaction}, we can see that the enhanced interaction offers consistent improvements over pronouns entities and also maintains the performance on other kinds of entities.

\subsection{Qualitative Analysis}
\begin{table*}[]
    \centering
    \small
    \begin{tabular}{c|l}
    \toprule
            1) Context & \multirow{2}{13cm}{\textbf{Today}, Taiwan is manifesting the elegance of a democratic island, once again attracting global attention, as the people on this land create a new page in our history.} \\
        & \\
        Groundtruth & \textit{time, date, day, today, present} \\
        Prediction & \textbf{Baseline}: \{\textcolor{red}{day$^\dagger$}, \textcolor{blue}{person$^\star$}, organization, religion\} \textbf{Ours}: \{day\} \\ \midrule
        
        2) Context & \multirow{2}{13cm}{\textbf{A gigantic robot} emerges, emitting a sound that paralyzes humans and disrupts all electrical systems in New York City.} \\
        & \\
        Groundtruth & \textit{object, device, machine, mechanism} \\
        Prediction & \textbf{Baseline}: \{object, \textcolor{red}{person$^\dagger$}, \textcolor{blue}{robot$^\star$}\} \textbf{Ours}: \{object, robot\}\\ \midrule
        
        3) Context & \multirow{2}{13cm}{\textbf{He} also has been accused of genocide in Bosnia and other war crimes in Croatia, but the date to try those two indictments together has not been set.}\\
        & \\
        Groundtruth & \textit{person} \\ 
        Prediction & \textbf{Baseline}:\{person, \textcolor{red}{god$^\dagger$}, title, \textcolor{blue}{criminal$^\star$}\} \textbf{Ours}: \{person, politician, criminal, male, prisoner\}\\ \midrule

        4) Context & \multirow{1}{13cm}{\textbf{Her} status was uncertain for Wimbledon, which begins June 23.} \\
        Groundtruth & \textit{person, athlete, adult, player, professional, tennis player, contestant} \\
        Prediction & \textbf{Baseline}: \{person, female, woman, spouse\} \textbf{Ours}: \{person, artist, female, woman\} \\ \midrule 
        
        5) Context & For eight years \textbf{he} treated thousands of wounded soldiers of the armed forces led by the CPC. \\
        Groundtruth & \textit{person, doctor, caretaker, nurse} \\
        Prediction & \textbf{Baseline}: \{person, soldier, suspect, serviceman\} \textbf{Ours}: \{person, soldier, man\}
        \\ \bottomrule
    \end{tabular}
    \caption{Qualitative analysis of validation samples.  We use different colors and subscript symbols to mark inconsistencies. The bottom two rows show error cases for both models.}
    \label{tab:human_analysis}
\end{table*}
 To gain insights on the improvements provided by our model, we manually analyze 100 error cases\footnote{The baseline model achieves the lowest precision on these 100 samples.} of the baseline model (\citet{choi2018ultra} with threshold 0.5) and see if our model can generate high-quality predictions. We first observe that many errors actually results from incomplete annotations. This suggests models' precision scores are often underestimated in this dataset. We discuss several typical error cases shown in Table~\ref{tab:human_analysis} and list more samples in the appendix (Table~\ref{tab:more_analysis}). 

A key observation is that while the baseline model tends to make inconsistent predictions (see examples 1, 2, 3), our model can avoid predicting such inconsistent type pairs. This indeed validates our model's ability to encode label correlations. We also notice that our model is more sensitive to gender information indicated by pronouns, while the baseline model sometimes holds the gender-indicating predictions and predict other types, our model predicts the gender-indicating types more often (examples 3, 4, 5). We conjecture that our model learns this easy way to maintain precision. 

For cases that both models fail, some of them actually require background knowledge (example 4) to make accurate predictions. Another typical case is that both models predict some other entities in the context (example 5). We think this potentially results from the data bias introduced by the head-word supervision. 

\begin{table}
    \centering
    \small
    \begin{tabular}{c|ccc}
    \toprule
        \textbf{Model} &  \textbf{Accuracy} & \textbf{Macro-F1} & \textbf{Micro-F1}\\ 
        \midrule
        AttentiveNER & 51.7 & 71.0 & 64.9 \\
        AFET & 55.1 & 71.1 & 64.7\\
        LNR & 57.2 & 71.5 & 66.1\\
        NFETC & \textbf{60.2} & 76.4 & 70.2 \\
        \midrule
        \citet{choi2018ultra} & 59.5 &  76.8 & 71.8\\
        \textsc{LabelGCN} & 59.6 & \textbf{77.8} & \textbf{72.2}\\
        \bottomrule
    \end{tabular}
    \caption{Results on OntoNotes. Upper rows show the results of baselines that explicitly use the hierarchical type structures.}
    \label{tab:onto}
\end{table}

\subsection{Evaluation on OntoNotes}
To better understand the requirements for applying our model, we further evaluate on the OntoNotes dataset. Here we do not apply the proposed mention-context matching module as this dataset does not include any pronoun entities. To obtain more reliable co-occurrence statistics, we use the augmented training data released by \citet{choi2018ultra}. However, since the training set is still much smaller than that of the Ultra-Fine dataset, the derived co-occurrence statistics are relatively noisy and might introduce undesired bias. We thus add an additional residual connection to our graph convolution layer, which allows the model to selectively use co-occurrence statistics. This indeed gives us improvements over previous state-of-the-arts, as shown in Table~\ref{tab:onto}. However, compared to Ultra-Fine, the margin of the improvement is smaller. In view of the key differences of these two datasets, we highlight two key requirements for our proposed model to offer substantial improvements. First, there should be a large-scale training set so that the derived co-occurrence statistics can reasonably reflect the true label correlations. Second, the samples themselves should also have higher label multiplicity. In fact, most of the samples in OntoNotes only have 1 or 2 labels. This property actually alleviates the need for models to capture label dependencies. 

\section{Conclusion}
In this paper, we present an effective method to impose label-relational inductive bias on fine-grained entity typing models. Specifically, we utilize a graph convolution layer to incorporate type co-occurrence statistics and word-level type similarities. This layer implicitly captures the label correlations in the latent vector space. Along with an attention-based mention-context matching module, we achieve significant improvements over previous methods on a large-scale dataset. As our method does not require external knowledge about the label structures, we believe our method is general enough and has the potential to be applied to other multi-label tasks with plain-text labels. 

\section*{Acknowledgement}
This research was supported in part by DARPA Grant D18AP00044 funded under the DARPA YFA program. The authors are solely responsible for the contents of the paper, and the opinions expressed in this publication do not reflect those of the funding agencies.

\bibliography{naaclhlt2019}
\bibliographystyle{acl_natbib}

\appendix
\clearpage
\onecolumn
\section{Appendix}
\begin{figure*}[h]
    \centering
    \includegraphics[width=\linewidth]{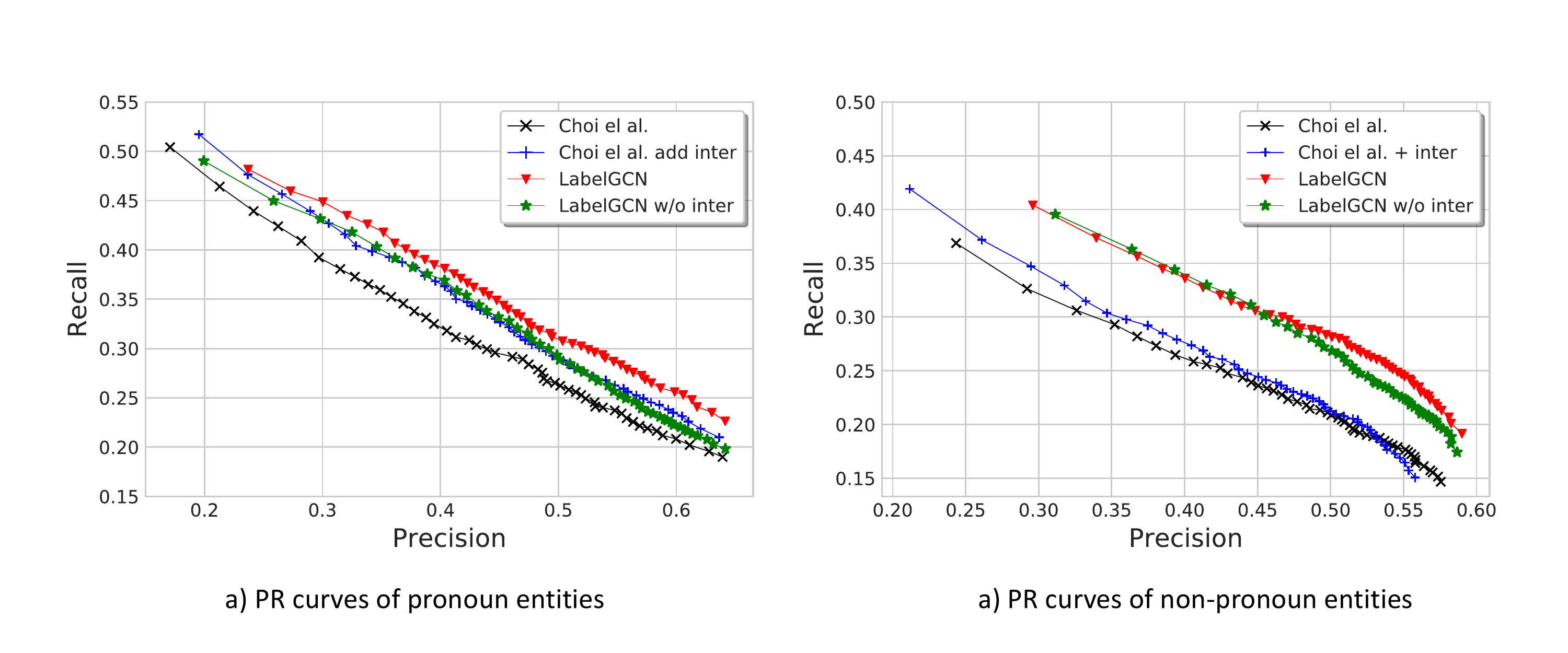}
    \caption{Precision-recall curves showing the decomposed results on pronoun and non-pronoun entity mentions. The enhanced mention-context interaction can consistently offer improvements for pronoun entity mentions while maintaining the performance for non-pronoun entity mentions.}
    \label{fig:more}
\end{figure*}

\begin{table*}[h]
\small
    \centering
    \begin{tabular}{c|c}
    \toprule
        learning rate & 0.001 \\
        batch size & 1000 \\
        position embedding size & 50 \\
        dropout on context $\mathcal{C}$ & 0.2 \\
        dropout on mention $\mathcal{M}$ & 0.5 \\
        hidden dimension of LSTM & 100 \\ 
        dropout on fused feature $f$ (Ultra-Fine) & 0.2 \\
        dropout on fused feature $f$ (OntoNotes) & 0.3\\
    \bottomrule
    \end{tabular}
    \caption{Hyperparameters used in our experiments}
    \label{tab:hyperparamters}
\end{table*}
\begin{table*}[t]
    \centering
    \small
    \begin{tabular}{c|l}
    \toprule
        
        Context & \multirow{1}{13cm}{\textbf{They} have been asked to appear in court to face the charge on Feb. 3.}\\
        Groundtruth & \textit{person, defendant, suspect, accused} \\ 
        Prediction & \textbf{Baseline}:\{person, engineer, officer, \textcolor{red}{policeman$^\dagger$}, \textcolor{blue}{prisoner$^\star$}, married, serviceman\} \textbf{Ours}:\{person\}\\  \midrule
        
        Context & \multirow{1}{13cm}{``\textbf{It} is truly a war crime," she added.} \\
        Groundtruth & \textit{event, crime, issue, offense, transgression, atrocity} \\
        Prediction & \textbf{Baseline}: \{\textcolor{red}{internet$^\dagger$}, event, \textcolor{blue}{art$^\star$}, writing\} \textbf{Ours}: \{law\}\\ \midrule
        
         Context & \multirow{2}{13cm}{\textbf{She} added that Israeli military personnel had conducted a medical examination after the shooting in concert with Palestinian medics.} \\
        & \\
        Groundtruth & \textit{person} \\
        Prediction & \textbf{Baseline}: \{\textcolor{red}{person$^\dagger$}, \textcolor{blue}{art$^\star$}, writing, convict, felon\} \textbf{Ours}: \{person, female, woman\} \\ \midrule

        Context & \multirow{2}{13cm}{The monument is located in Pioneer Park Cemetery in the Convention Center District of downtown Dallas, Texas, \textbf{USA}, next to the Dallas Convention Center and Pioneer Plaza.}\\
        & \\
        Groundtruth & \textit{location, place, country, area, nation, region}\\
        Prediction & \multirow{2}{13cm}{\textbf{Baseline}: \{\textcolor{red}{location$^\dagger$}, \textcolor{blue}{person$^\star$}, agency, artist, cemetery, country, language, title, republic\} \textbf{Ours}:\{nationality, location, place, country, area, license, nation\}}\\ 
        & \\ \midrule
        
        Context & \textbf{The committee} undertook its work on Saturday 16/2/1426 A. H . The following is noteworthy : \\
        Groundtruth & \textit{group, organization, agency, company, institution, administration, body, management, party}\\
        Prediction & \textbf{Baseline}: \{\textcolor{red}{committee$^\dagger$}, \textcolor{blue}{person$^\star$}, organization, government\} \textbf{Ours}: \{group, government, committee\}
        
        \\ \midrule
        Context & \multirow{2}{13cm}{\textbf{They} are accused of helping Libya develop a nuclear weapons programme and were alleged to have been in contact with Abdul Qadeer Khan , the disgraced father of Pakistan 's nuclear programme.} \\
        & \\
        Groundtruth & \textit{group, terrorist} \\
        Prediction & \textbf{Baseline}: \{military, \textcolor{red}{person$^\dagger$}, group, \textcolor{blue}{country$^\star$}\} \textbf{Ours}: \{person, politician, prisoner, serviceman\}
        
        \\ \midrule
        Context & \multirow{2}{13cm}{\textbf{It} also marked the first major roundup of Islamist leaders by a government eager to demonstrate its commitment to the anti-terror fight waged by the United States.}\\
        & \\
        Groundtruth & \textit{event, consequence} \\ 
        Prediction & \textbf{Baseline}:\{\textcolor{red}{internet$^\dagger$}, event, \textcolor{blue}{art$^\star$}, writing\} \textbf{Ours}:\{event\} \\  \midrule
        
        Context & \multirow{2}{13cm}{If you have ever watched a keynote speech by \textbf{Steve Jobs}, you know that he was the best of the best in launching a product.} \\
        & \\
        Groundtruth & \textit{person, adult, businessman, celebrity, professional} \\
        Prediction & \textbf{Baseline}: \{person, artist, \textcolor{red}{athlete$^\dagger$}, author, \textcolor{blue}{musician$^\star$}\} \textbf{Ours}:\{person\} \\ \midrule
        
        Context & \multirow{2}{13cm}{\textbf{They} dined together,  this time in Benedict's house, before the pope was driven back to his temporary residence in Regensburg 's St Wolfgang Seminary.} \\
        & \\
        Groundtruth & \textit{adult, man, supporter, serviceman}\\
        Prediction & \textbf{Baseline}: \{person, adult, female, woman\} \textbf{Ours}: \{person\} \\ \midrule

        Context & \multirow{1}{13cm}{Topic : \textbf{I} am grateful to the University of Science and Technology}\\
        Groundtruth & \textit{person, individual, student} \\ 
        Prediction & \textbf{Baseline}:\{person, politician, employee, leader, minister, traveler, announcer, clergyman\} \textbf{Ours}:\{person, student\}\\ \midrule

        Context & \multirow{1}{13cm}{``\textbf{I} didn't think the speech was that long,'' Pataki said.} \\
        Groundtruth & \textit{person, speaker} \\
        Prediction & \textbf{Baseline}: \{person, actor, politician, spokesperson, woman\} \textbf{Ours}: \{person, adult\} \\ \midrule
        
        Context &  "This is touching our troops," \textbf{she} said.\\
        Groundtruth & \textit{person, adult, female, reporter, woman}\\
        Prediction & \multirow{2}{13cm}{\textbf{Baseline}: \{person, politician, official, spokesperson, communicator\} \textbf{Ours}: \{female, official, reporter, strategist, communicator, officeholder\}}\\ & \\

        \bottomrule
    \end{tabular}
    \caption{More sample predictions. Our model is able to give more accurate type predictions and also reduce the inconsistency in the output type set.}
    \label{tab:more_analysis}
\end{table*}

\end{document}